# Physics-Guided VLM Priors for All-Cloud Removal

Liying Xu, Huifang Li, Huanfeng Shen

*Abstract*—Cloud removal is a fundamental challenge in optical remote sensing due to the heterogeneous degradation. Thin clouds distort radiometry via partial transmission, while thick clouds occlude the surface. Existing pipelines separate thin-cloud correction from thick-cloud reconstruction, requiring explicit cloud-type decisions and often leading to error accumulation and discontinuities in mixed-cloud scenes. Therefore, a novel approach named Physical-VLM All-Cloud Removal (PhyVLM-CR) that integrates the semantic capability of Vision-Language Model (VLM) into a physical restoration model, achieving high-fidelity unified cloud removal. Specifically, the cognitive prior from a VLM (e.g., Qwen) is transformed into physical scattering parameters and a hallucination confidence map. Leveraging this confidence map as a continuous soft gate, our method achieves a unified restoration via adaptive weighting: it prioritizes physical inversion in high-transmission regions to preserve radiometric fidelity, while seamlessly transitioning to temporal reference reconstruction in low-confidence occluded areas. This mechanism eliminates the need for explicit boundary delineation, ensuring a coherent removal across heterogeneous cloud covers. Experiments on real-world Sentinel-2 surface reflectance imagery confirm that our approach achieves a remarkable balance between cloud removal and content preservation, delivering hallucination-free results with substantially improved quantitative accuracy compared to existing methods.

## I. INTRODUCTION

CLOUD contamination remains a persistent bottleneck in optical remote sensing. Crucially, cloud optical thickness varies continuously within scenes. As optical thickness increases, the dominant degradation shifts from radiometric distortion caused by thin-cloud partial scattering to total information loss caused by thick-cloud occlusion. This spatial continuity renders radiometric correction and information reconstruction inseparable, fusing them into a single ill-posed problem where no distinct boundary exists.

Given the distinct optical behaviors of different cloud types, existing literature typically addresses cloud removal through two specialized streams[1], [2]. Thin clouds are usually corrected via radiometry-aware inversion of atmospheric scattering models[3], utilizing methods such as the Dark Channel Prior (DCP)[4]or Haze-Optimized Transformation (HOT)[5]. Conversely, thick clouds are handled as information reconstruction using auxiliary observations[6], relying on multi-temporal reconstruction[7] or generative synthesis[8] to recover missing content. Despite the progress of deep learning in both settings, model training and architectural choices are still shaped by these divergent physical assumptions, which hinders the development of a unified framework. The resulting reliance on cloud detection makes removal quality heavily contingent on segmentation accuracy. Misalignment in transition zones inevitably leads to error propagation and visible boundary artifacts.

Therefore, a unified all-cloud removal framework named Physical-VLM All-Cloud Removal (PhyVLM-CR) is proposed in this letter that synergizes the semantic perception of Vision-Language Model (VLM) with the rigorous constraints of physical radiometry. The VLM is redefined as a cognitive prior extractor, rather than a pixel generator, to guide the derivation of scattering parameters and a hallucination confidence map. These scattering parameters enable global thin cloud removal, while the confidence map functions as a continuous soft gate to adaptively integrate the temporal reconstruction content in occluded regions. In summary, this letter makes three main contributions.

1) **A unified zero-shot all-cloud removal method** that obviates explicit classification, preserving the spatial continuity of cloud degradation.

2) **A cognitive prior extraction strategy** that utilizes VLM semantics to guide the derivation of scattering parameters and a confidence map for hallucination suppression.

3) **An adaptive fusion mechanism** that seamlessly integrates physical inversion for thin clouds with temporal reconstruction for thick clouds, ensuring coherent removal in mixed cloud scenes.

## II. MODEL

To characterize the formation of cloud-degraded imagery resulting from the interaction of radiation with the land surface and turbid atmosphere, the imaging model is formulated as:

$$I(x) = J(x)t(x) + A(1-t(x)) \qquad (1)$$

where, $x$ denotes the pixel coordinate, $I(x)$ represents the observed cloudy image, and $J(x)$ denotes the target cloud-free surface reflectance. The transmission map $t(x) \in [0,1]$ quantifies the portion of surface radiance that effectively reaches the sensor, while $A$ represents the global atmospheric light.

Through large-scale pre-training, VLM possesses strong semantic reasoning capabilities to infer latent scene structures and textures from cloud degraded observations. Nevertheless,

Liying Xu, Huifang Li are with School of Resource and Environmental Sciences, Wuhan University, Wuhan 430079, China (e-mail: liyingxuwhu@whu.edu.cn; huifangli@whu.edu.cn).

Huanfeng Shen is with the School of Resource and Environmental Sciences and the Collaborative Innovation Center for Geospatial Technology, Wuhan University, Wuhan 430079, China (e-mail: shenhf@whu.edu.cn).

such data-driven inference is inherently radiometrically unreliable and may introduce generative artifacts. To account for this uncertainty, we model the true surface reflectance as the VLM prediction corrected by a residual term:

$$J(x) = J_{VLM}(x) + E(x) \qquad (2)$$

where, $J_{VLM}(x)$ denotes the VLM-generated prediction, and $E(x)$ denotes the hallucination residual, capturing the discrepancy between the generative prediction and the authentic radiometric reality.

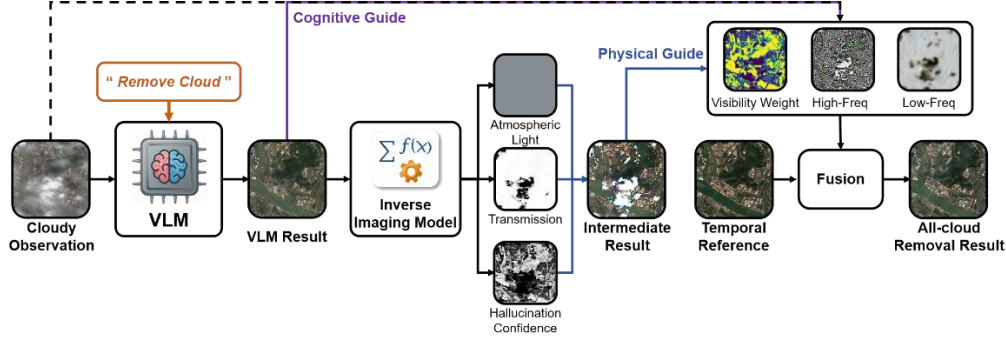

**Fig. 1.** Overview of the proposed PhyVLM-CR.

### III. METHOD

The proposed method is shown in Fig. 1. The proposed framework integrates semantic perception with physical constraints through three sequential stages: cognitive prior acquisition, physics-guided parameter extraction, and unified all-cloud removal.

*A. Cognitive Prior Acquisition*

The framework leverages the semantic reasoning capacity of a large-scale pre-trained VLM and its associated image-editing backbone. Specifically, Qwen-Image-Edit[9] is employed to produce an initial all-cloud removal candidate $J_{VLM}(x)$ from the degraded observation $I(x)$ driven by the prompt "remove cloud" in Chinese.

Attributable to the human-like cognitive perception of VLM, $J_{VLM}(x)$ exhibits plausible scene structures and global illumination contexts. However, lacking rigorous physical constraints, the generative predictions are prone to hallucinations and spurious artifacts. Consequently, $J_{VLM}(x)$ is utilized strictly as a carrier of cognitive priors to guide parameter estimation, rather than as the final cloud removal result.

*B. Physics-guided Parameter Extraction*

To convert the cognitive prior into physically grounded constraints, the global atmospheric light $A$, transmission map $t(x)$, and hallucination confidence map $U(x)$ are estimated by fitting a scattering model to the observation $I(x)$ under the guidance of the prior carrier $J_{VLM}(x)$.

Due to the mixed cloud conditions, the global atmospheric light $A$ is regressed from the image region exhibiting the highest cloud probability. The candidate region $\Omega$ is identified via a probability map combining high brightness, low saturation, and minimal texture gradient to ensure a robust capture of the ambient veil color. The value of $A$ is derived as the spatial median of $\Omega$:

$$A = \operatorname{median}_{x \in \Omega}(I(x)) \qquad (3)$$

$$\Omega = \{x \mid \sigma(V_x) \cdot \sigma(1 - S_x) \cdot \sigma(-\|\nabla I_x\|) > \kappa\} \qquad (4)$$

where, $V_x$, $S_x$, and $\|\nabla I_x\|$ denote the brightness, saturation, and gradient magnitude at pixel $x$, respectively. The sigmoid functions $\sigma$ effectively gate pixels that exhibit high brightness, low saturation, and low texture. The threshold $\kappa$ is set to the 85% of the probability map to ensure $A$ is statistically derived solely from the purest cloud regions.

With the atmospheric light $A$ anchored, the transmission map $t(x)$ is estimated by initially assuming the residual error $E(x)$ to be negligible. This simplification reduces the ill-posed problem to a tractable regression between the observed cloudy image $I(x)$ and the VLM prediction $J_{VLM}(x)$. To suppress local hallucination misalignment, the images are decomposed into base layers, denoted by the superscript $B$, using edge-preserving filtering. The transmission $t(x)$ is then derived via a robust pixel-wise estimation:

$$t(x) = \frac{(I^B(x) - A)^\top (J_{VLM}^B(x) - A)}{\|J_{VLM}^B(x) - A\|^2 + \varepsilon} \qquad (5)$$

where, $\varepsilon = 10^{-6}$ prevents numerical instability.

To compensate for potential deviations from the zero-error assumption, a hallucination confidence map $U(x)$ is formulated to quantify the magnitude of the residual term $E(x)$ within the scattering model. A frequency-decoupled strategy is adopted to distinguish global physical inconsistencies from local hallucination misalignments. The global physical inconsistencies $r(x)$ are first calculated within the edge-preserving base layer spaces:

$$r(x) = \left\| I^B(x) - \left( t(x) J_{VLM}^B(x) + (1 - t(x)) A \right) \right\|_2 \qquad (6)$$

Conversely, to capture local hallucination misalignment that are smoothed out in the base layers, the final confidence $U(x)$ incorporates a high-frequency penalty on the raw intensity domain:

$$U(x) = \exp\left(-\frac{r(x)}{\lambda_{phy}}\right) \cdot \exp\left(-\frac{\max\left(0, \mathrm{H}\left(J_{VLM}(x)\right) - \mathrm{H}\left(I(x)\right)\right)}{\lambda_{hall}}\right) \quad (7)$$

where, $\mathrm{H}(\cdot)$ extracts high-frequency texture intensity by Laplacian operator. The normalization parameters $\lambda_{phy}$ and $\lambda_{hall}$ are adaptive, set to the 75% of the re-synthesized residual map and the high-frequency difference map, respectively.

To strictly preserve edges and suppress noise, the transmission map $t(x)$ is refined using a guided filter weighted by the hallucination confidence map $U(x)$.

*C. Unified All-cloud Removal*

With the physical scattering parameters and hallucination confidence map determined, the unified all-cloud removal aims to integrate the physical fidelity of the scattering model with the semantic reasoning capability of the VLM. The method is designed to progressively recover the surface radiance through physical inversion, cognitive adjustment, and occlusion reconstruction. A preliminary physical estimate $J_{phy}(x)$ is first obtained by inverting the imaging model:

$$J_{phy}(x) = \frac{I(x) - A}{\max(t(x), t_0)} + A \quad (8)$$

where, $t_0$ is a lower bound to prevent numerical instability in low-transmission regions.

However, due to the ill-posed nature of single-image dehazing, $J_{phy}(x)$ often exhibits color distortion and insufficient contrast. Consequently, cognitive priors from the VLM are required to optimize the radiometric quality.

To rectify these distortions without introducing generative artifacts, a frequency-decoupled cognitive adjustment is formulated. This strategy separates cognitive correction from textural preservation using a low-pass filter $lp$. The adjusted image $J_{cog}(x)$ is synthesized by integrating cognitive correction and high-frequency detail injection into the physical baseline:

$$\begin{aligned} J_{cog}(x) &= J_{phy}(x) \\ &+ \alpha U(x)\left(J_{VLM}^{lp}(x) - J_{phy}^{lp}(x)\right) \\ &+ \beta t(x)\left(I(x) - I^{lp}(x)\right) \end{aligned} \quad (9)$$

where, the scalar coefficients $\alpha$ and $\beta$ modulate the intensity of cognitive guidance and detail preservation, respectively. This design ensures the all-cloud removal result possesses natural color representation while faithfully preserving the sensor-captured structure, effectively discarding uncontrollable high-frequency hallucinations from the VLM.

In regions where clouds completely obstruct the surface signal, rendering physical correction intractable, an information reconstruction mechanism is triggered. An adjacent temporal reference $I_{ref}$ is radiometrically aligned to the current scene using linear parameters estimated from high-visibility areas. The final output $J(x)$ fuses the cognitively adjusted image with the reference via a visibility weight $\omega(x)$:

$$J(x) = \omega(x)J_{cog}(x) + (1 - \omega(x))I_{ref}(x) \quad (10)$$
$$\omega(x) = \exp(-\gamma(1 - t(x))) \quad (11)$$

Inspired by the radiative transfer principles utilized in atmospheric correction models[10], the parameter $\gamma$ is set to 4.0. This value functions as an extinction sensitivity coefficient, simulating an effective optical depth limit where the surface signal transmission drops below 2% ($e^{-4} \approx 0.018$). This ensures that regions covered by thick cloud are rigorously constrained to rely on the temporal reference, suppressing potential artifacts from the physical inversion and the generative hallucination from VLM.

IV. EXPERIMENT

*A. Data*

This study conducts experimental validation on Sentinel-2 surface reflectance imagery with heterogeneous cloud cover. Details are provided in the comparative experiments section.

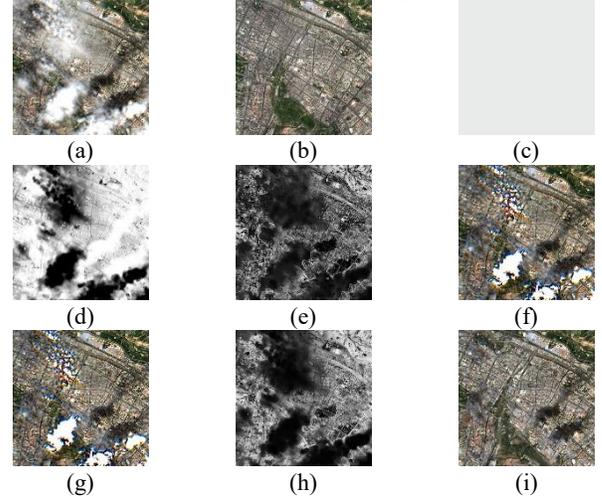

**Fig. 2.** The intermediate results of the proposed PhyVLM-CR. (a) Cloudy image $I(x)$. (b) Cloud removal result of VLM $J_{VLM}(x)$. (c) Atmospheric light $A$. (d) Transmission $t(x)$. (e) Hallucination confidence $U(x)$. (f) Physical estimate result $J_{phy}(x)$. (g) Cognitive adjusted result $J_{cog}(x)$. (h) Visibility weight $\omega(x)$. (i) Cloud removal result of PhyVLM-CR $J(x)$.

*B. Process Validation*

Fig. 2 illustrates intermediate results under complex cloud conditions. The VLM output provides a globally coherent hypothesis but lacks fine-scale radiometric fidelity. Consequently, it is more appropriate to serve as a carrier of cognitive priors rather than a direct restoration result. Then, the physically interpretable variables are extracted from the cognitive prior, as shown in Figs. 2(c)-(e). Guided by these estimates, the $J_{phy}(x)$ preserves radiometric plausibility in high-transmission areas, while the $J_{cog}(x)$ suppresses VLM artifacts, as shown in Figs. 2(f)-(h). A visibility weight $\omega(x)$ ensures a smooth transition. Consequently, the final output $J(x)$ achieves spatially coherent removal across all cloud zones without relying on rigid segmentation.

*C. Comparison*

Quantitative and qualitative comparisons are conducted on complex cloud-degraded scenes using a paired multi-temporal protocol. For each scene, the clear-sky observation temporally nearest to the cloudy acquisition (Cloudy_$T_1$) is selected as the ground truth (Truth_$T_2$) for quantitative evaluation, while the second-nearest clear-sky observation (Ref_$T_3$) serves as an auxiliary reference for occlusion reconstruction. Detailed dataset specifications are provided in Table I. Note that the acquisition year is omitted from the table for brevity, as all data were collected in 2023. This setup enables a consistent evaluation of both radiometric correction and information reconstruction under real-world temporal availability constraints. Representative qualitative comparisons are shown in Fig. 3, with quantitative metrics summarized in Table II.

TABLE I
DETAILS OF THE DATASET USED IN EXPERIMENTS

| Scene | Center Coordinates | $T_1$ | $T_2$ | $T_3$ |
|---|---|---|---|---|
| Sichuan | [30.60, 105.50] | 04-01 | 04-11 | 05-06 |
| Hainan | [18.25, 109.51] | 10-04 | 11-18 | 11-23 |
| Qinghai | [36.62, 101.78] | 07-16 | 07-21 | 06-26 |
| Hubei | [30.55, 114.37] | 10-06 | 10-16 | 10-31 |
| Jiangsu | [31.50, 120.50] | 10-05 | 10-15 | 11-14 |
| Yunnan | [25.04, 102.70] | 01-11 | 01-26 | 01-31 |

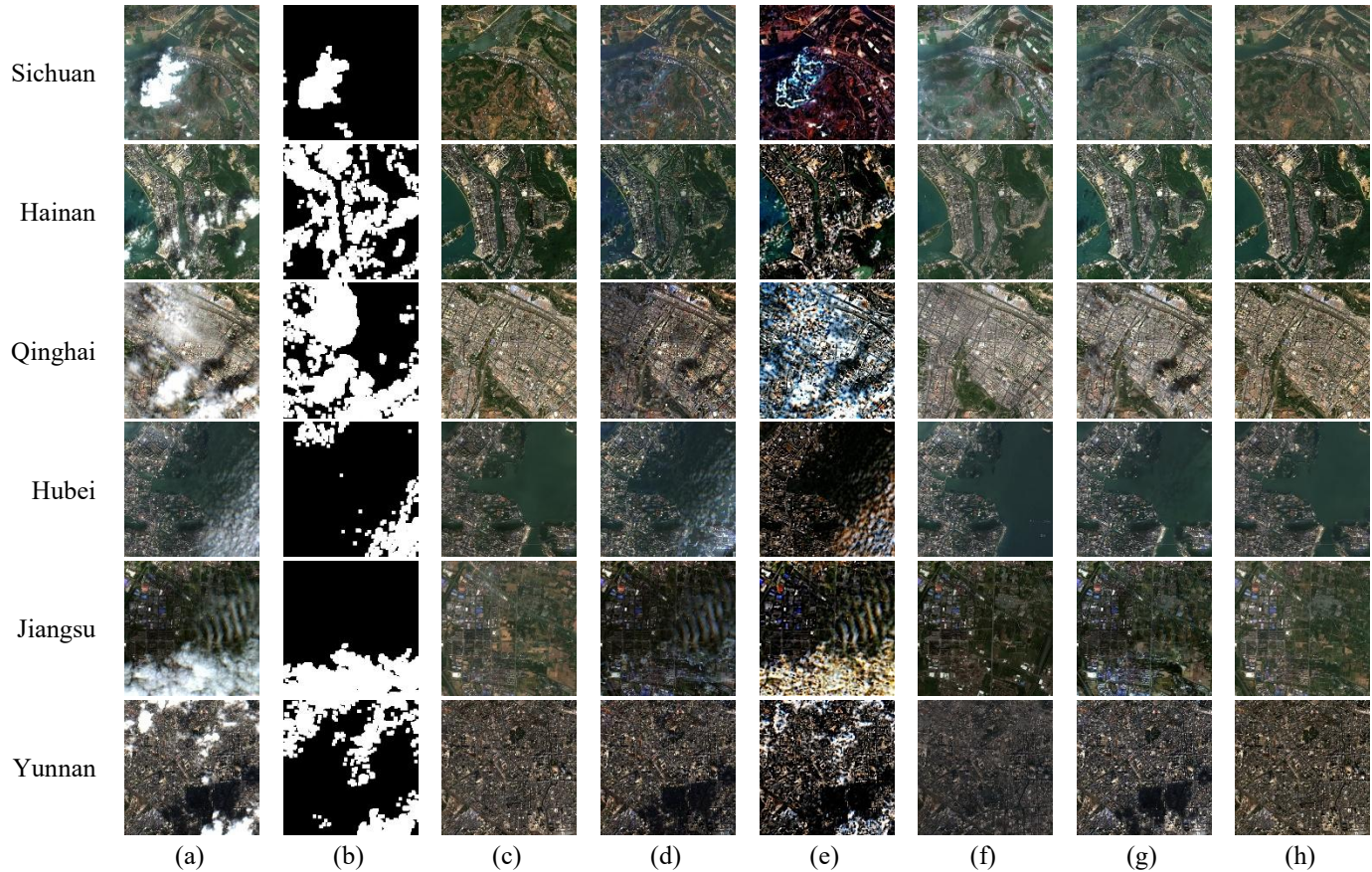

**Fig. 3.** All-cloud removal results based on the various methods. (a) Real cloudy images acquired in $T_1$. (b) Thick-cloud masks calculated from MSK_CLDPRB. (c) Real clear images acquired in $T_3$ for reference. (d) Traditional physical method. (e) Zero-shot deep learning method. (f) Vision-Language Model. (g) The proposed PhyVLM-CR. (h) Real clear images acquired in $T_2$ as ground truth.

TABLE II
EVALUATION FOR THE CLOUD REMOVAL

| Scene | **Traditional Physical** | | **Zero-shot Deep Learning** | | **Vision-Language Model** | | **PhyVLM-CR** | |
|---|---|---|---|---|---|---|---|---|
| | PSNR↑ | SSIM↑ | PSNR↑ | SSIM↑ | PSNR↑ | SSIM↑ | PSNR↑ | SSIM↑ |
| Sichuan | 17.583 | 0.5492 | 12.450 | 0.2679 | <u>20.890</u> | <u>0.5706</u> | **22.562** | **0.7937** |
| Hainan | <u>17.531</u> | <u>0.5556</u> | 12.737 | 0.2801 | 13.995 | 0.2547 | **19.155** | **0.7294** |
| Qinghai | <u>15.847</u> | <u>0.7698</u> | 9.708 | 0.3022 | 14.456 | 0.3182 | **18.771** | **0.8256** |
| Hubei | <u>19.701</u> | <u>0.7313</u> | 14.568 | 0.3042 | 18.937 | 0.5655 | **27.188** | **0.9220** |
| Jiangsu | 15.495 | <u>0.5258</u> | 11.073 | 0.2211 | <u>18.324</u> | 0.4125 | **19.904** | **0.7201** |
| Yunnan | <u>17.214</u> | <u>0.6522</u> | 11.512 | 0.3106 | 13.229 | 0.0323 | **19.865** | **0.7888** |

As illustrated in Fig. 3(b), thick-cloud masks are derived from the official Sentinel-2 cloud probability product (MSK_CLDPRB) with a threshold of 40. To address the resolution mismatch between the 10 m visible bands and the 20 m probability product, the masks were resampled and then refined using morphological closing and opening. These operations served to remove scattered pixels and consolidate cloud regions, respectively. These processed masks are then utilized to guide the cloud removal workflow for the first two comparative methods.

The first physical compared method employs a traditional strategy that explicitly separates thin-cloud correction from thick-cloud restoration. Specifically, thin-cloud regions are corrected using SSADCP[11], while thick-cloud occlusions are reconstructed using FRARC[12]. The results are presented in Fig. 3(c). Based on visual inspection, the mask-guided approach exhibits limitations at cloud boundaries, where imprecise transitions fail to correct thick-to-thin cloud zones, causing cloud residual. Simultaneously, bright surface features are prone to being misidentified as clouds that leads to the replacement of valid observations with temporal reference data.

The second compared method is a zero-shot Deep Learning pipeline designed to ensure fair comparison in the absence of mixed cloud paired training data. Thin-cloud correction adopts Zero-Shot Image Dehazing (ZID)[13], utilizing MSK_CLDPRB to guide transmission map estimation. Thick-cloud reconstruction is performed via Deep Image Prior (DIP)[14] in an inpainting configuration using the thick-cloud masks. In the experimental setup, both ZID and DIP are configured with 500 iterations, resulting in a processing time of approximately 30 minutes per scene and significantly exceeding the second-level efficiency of other methods. Despite the high computational cost, performance remains unsatisfactory due to the domain gap between natural images and remote sensing data. Specifically, networks designed to remove homogeneous haze in natural images struggle to handle the heterogeneous cloud patterns and complex ground structures to remote sensing data.

The third compared method evaluates a purely VLM generative solution by directly reporting the output of Qwen-Image-Edit[9] under the "remove cloud" textual prompt. In the absence of physical constraints, the VLM relies solely on semantic priors, leading to severe hallucinations where valid ground details are replaced by fictitious textures, including non-existent land-cover patterns and even spurious characters, particularly in thick-cloud regions. This misalignment with actual ground surfaces leads to inferior quantitative performance, highlighting the risks of directly applying VLM to remote sensing data.

The proposed method strategically integrates the cognitive prior of VLMs with the rigorous constraints of physical model. Effectively mitigate the hallucination risks associated with purely generative approaches while overcoming the domain limitations of traditional zero-shot deep learning methods. Consequently, the proposed PhyVLM-CR recovers accurate surface details and maintains spectral fidelity even in complex transition zones, achieving the superior visual quality and quantitative accuracy.

## V. Conclusion

This letter presents PhyVLM-CR, a unified VLM-guided, physics-constrained framework tailored for all-cloud removal in remote sensing imagery. Departing from traditional approaches, PhyVLM-CR redefines the role of generative artificial intelligence, treating the VLM output not as a final result but as a cognitive prior to derive physics-interpretable components. This design allows for the radiometrically accurate removal of thin clouds via physical inversion, while seamlessly reconstructing thick-cloud occlusions by temporal references. Crucially, unlike traditional strategies that rely heavily on binary cloud detection, the proposed PhyVLM-CR implements a continuous fusion mechanism regulated by these physical priors. This design eliminates hard boundaries between cloud types, delivering a coherent all-cloud removal result.